\documentclass[10pt,journal,compsoc]{IEEEtran}
\ifCLASSOPTIONcompsoc
  \usepackage[nocompress]{cite}
\else
  \usepackage{cite}
\fi
\ifCLASSINFOpdf
\else
\fi

\usepackage{times}
\usepackage{epsfig}
\usepackage{graphicx}
\usepackage{amsmath}
\usepackage{amssymb}
\usepackage{caption}
\usepackage{subcaption}
\usepackage{booktabs}

\usepackage[pagebackref=true,breaklinks=true,letterpaper=true,colorlinks,bookmarks=false]{hyperref}

\def\RR{\mathbb{R}}

\def\mat#1{{\mathcal{#1}}}
\def\vect#1{\mbox{\boldmath $#1$}}

\begin{document}
%
\title{One shot Joint Colocalization \& Cosegmentation}


\author{\IEEEauthorblockN{Abhishek Sharma},
\IEEEauthorblockA{Max Planck Institute for Informatics \\
                        asharma@mpi-inf.mpg.de}
}


%



\IEEEtitleabstractindextext{%
\begin{abstract}
 This paper presents a novel framework in which image cosegmentation and colocalization are cast into a single optimization problem that integrates information from low level appearance cues with that of high level localization cues in a very weakly supervised manner. In contrast to multi-task learning paradigm that learns similar tasks using a shared representation, the proposed framework leverages two representations at different levels and  simultaneously discriminates between foreground and background at the bounding box and superpixel level using discriminative clustering. We show empirically that constraining the two problems at different scales enables the transfer of semantic localization cues to improve cosegmentation output whereas local appearance based segmentation cues help colocalization. The unified framework outperforms strong baseline approaches, of learning the two problems separately, by a large margin on four benchmark datasets. Furthermore, it obtains competitive results compared to the state of the art for cosegmentation on two benchmark datasets and second best result for colocalization on Pascal VOC 2007. 
\end{abstract}
\begin{IEEEkeywords}
Discriminative clustering, weak supervision, cosegmentation, colocalization, multi-task learning
\end{IEEEkeywords}
}

\maketitle
\IEEEdisplaynontitleabstractindextext
%
\IEEEpeerreviewmaketitle

\section{Introduction}
\label{Intro}
Localizing and segmenting objects in an image is a fundamental problem in computer vision since it facilitates  many high level vision tasks such as object recognition, action recognition~\cite{Yang2010}, natural language description of images~\cite{Karpathy14} to name a few. Thus, any advancements in image segmentation and localization algorithm are automatically transferred to the performance of high level tasks~\cite{Karpathy14}. \\
\par With the recent success of deep networks, supervised top down segmentation methods obtain impressive performance~\cite{Long2015} by learning on pixel level labelled datasets. The same is true for object detection~\cite{ross14rcnn}. However, the amount of annotations required to achieve pixel or bounding box  labelled datasets is tremendous~\cite{Joulin15}. Taking into account the cost of obtaining such annotations, recent work has explored  the problem of weakly-supervised object discovery~\cite{Pandey11,RJKL13,Wang14,cho15}. The degree of supervision used in these problems
varies from weak ( positive and negative image-level labels for a target class~\cite{Siva13}), very weak
( image level labels e.g. colocalization ~\cite{DAF12,TJLF14} and cosegmentation~\cite{JBP10,Irani13}, and null~\cite{cho15}. In this paper, we focus on colocalization and cosegmentation and use very weak supervision to imply that labels are given only at the image level.\\
\par  Cosegmentation is the problem of segmenting common foreground regions out of a set of images whereas colocalization aims to localize the common object. Prior work in the supervised setting has used off-the-shelf object detectors to guide the segmentation process~\cite{Ladicky10} and also used segmentation as an initial phase for detection. However, existing work for cosegmentation and colocalization either completely ignores these complimentary cues  or use them in a two stage decision process, either as pre-processing step~\cite{Quan_16} or for post processing~\cite{Li16}. For example, Quan {\em et al.}~\cite{Quan_16}  refines the coarse localization heat map obtained by a VGG network~\cite{Chatfield14} to improve cosegmentation. However, it is difficult to recover from errors introduced in the initial stage and the post processing steps are prone to unwanted heuristics. \\

\par This paper advocates an alternative to the prevalent trends
of either ignoring these complimentary cues or placing a clear separation between segmentation and localization. In the weakly supervised scenario, the goal of knowledge transfer between the two tasks becomes even more challenging.  The key idea here is to avoid making hard decisions and instead, couple these two problems by linear constraints. We empirically show that constraining the two problems jointly improves the performance of both tasks significantly. Our work, although similar in spirit to the prior work that embeds pixels and parts in a graph ~\cite{Xu_02,MYP_11}, builds on the discriminative framework of ~\cite{MaxMar05,BaHa07} which utilizes a more powerful top down maximum margin machinery in an unsupervised fashion. \\
\par Contrary to the conventional approach of multi task learning.~\cite{MTL96,Lapin14} where two (or more) similar tasks are jointly learned using a shared representation, we instead leverage two representations at different scales and enable the transfer of information  implicit in these representations during a one shot optimization scheme. More precisely, the proposed formulation exploits the semantic, localization cues of bounding boxes to guide cosegmentation and leverages low level segmentation appearance cues  cues at superpixel level to improve colocalization. \\
  
\begin{figure*}
\centering
\vskip .15in
\includegraphics[scale = .5]{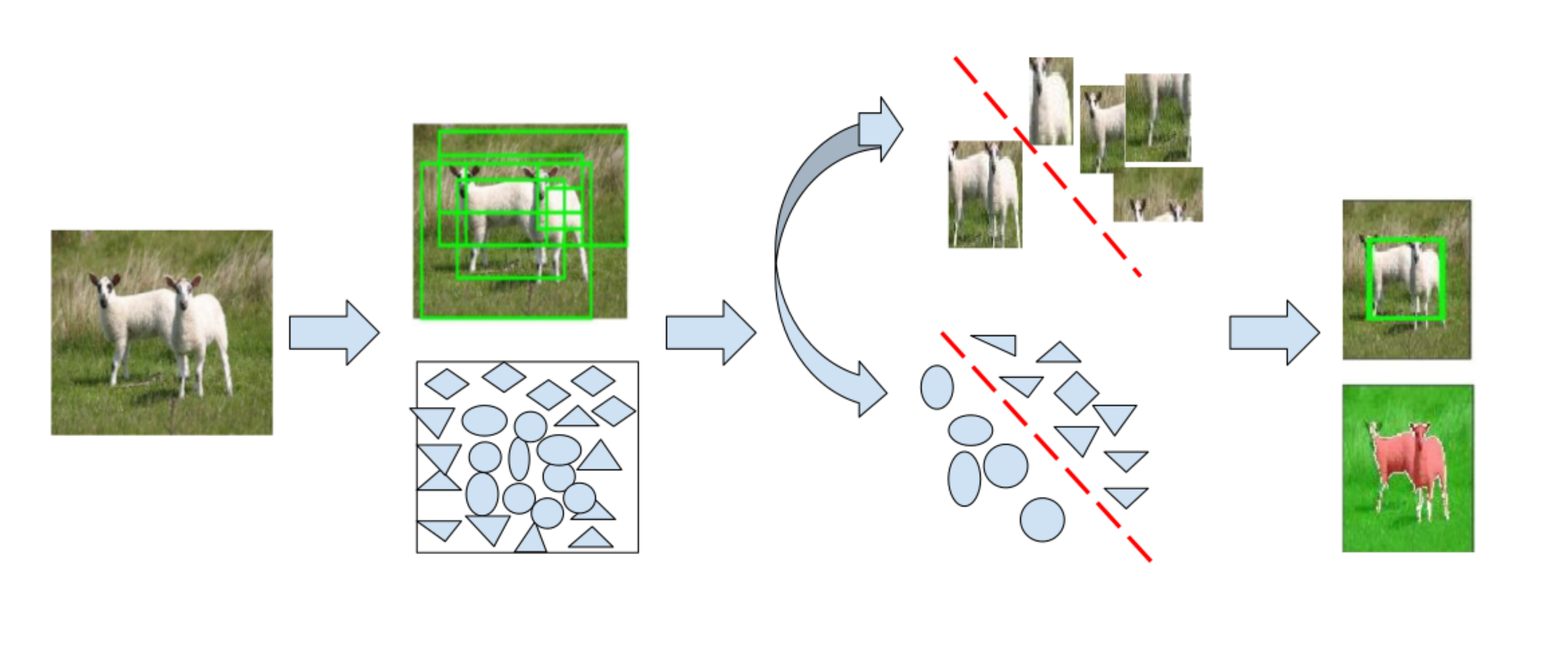}
\caption{Given an image, our framework generates bounding boxes (top) and features (bottom) for superpixel and bounding boxes. It then simultaneously learns to classify bounding boxes and  superpixels in foreground and background}
\vskip .15in
\end{figure*}
 
 Our contributions are as follows:  
  1) We propose a novel framework that simultaneously learns to localize and segment common objects in images. The unified framework obtains competitive results compared to the state of the art for cosegmentation on three benchmark datasets and second best result for colocalization on Pascal VOC 2007.  
     2) We show a novel mechanism to constrain the two problems via linear constraints in an unsupervised way that lifts the output performance of cosegmentation and  colocalization by more than 10 points.  
    3)We provide an extensive evaluation of our approach that shows contribution of novel terms in the objective function and the effect of different cues on three benchmark datasets.

\section{Related work}
 \par \noindent\textbf{Supervised Setting.} Numerous works have used off-the-shelf object detectors to guide segmentation process. Ladicky {\em et al.}~\cite{Ladicky10} used object detections as higher order potentials in a CRF-based segmentation system by encouraging all pixels in the foreground of a detected object to share the same category label as that of the detection.  Xu {\em et al.}~\cite{xu_15} also used bounding box as a weak supervision for semantic segmentation.Vicente {\em et al.}~\cite{VRK11}  introduced the idea of using bounding box for cosegmentation in a supervised setting. Alternatively, segmentation cues have been used before to help detection~\cite{Parkhi11,Li16}. Parkhi {\em et al.}~\cite{Parkhi11} uses color models from predefined rectangles on cat and dog faces to do GrabCut~\cite{GrabCut} and improve the predicted bounding box.
Hariharan {\em et al.}~\cite{hari_malik14} used CNN to simultaneously detect and segment by classifying image regions. All these approaches require ground truth annotation either in the form of bounding boxes or segmented objects during training phase which is challenging to obtain on large scale. \\
\par \noindent \textbf{Weakly Supervised Setting.}
Rother {\em et al.}~\cite{RKMB06} first introduced the idea of cosegmentation in a relatively simple setting where the same object lies in front of different backgrounds in a pair of images. Since then, many work~\cite{MSD09,JBP10, MSP11,RSLP12,WHG13,Quan_16} have been proposed to improve cosegmentation performance which can be broadly classified into discriminative and similarity based approaches. Similarity based approaches~\cite{RKMB06,VRK11,RSLP12, RJKL13} exploit the information of having common foreground across images and seek to segment it out by learning the foreground distribution or matching it across images~\cite{RJKL13,Irani13}. For example, Faktor \& Irani~\cite{Irani13} propose to discover the co-occurring regions first, and then perform cosegmentation by mapping between the co-occurring regions across the different images. In contrast, discriminative techniques~\cite{JBP10,JBP12} mainly rely on separating a set of images into most separable clusters while taking care of local spatial consistency. For example,Joulin {\em et al.}~\cite{JBP10} leverages discriminative clustering~\cite{BaHa07} to segment out the most discriminative parts in a set of images. However, most of these approaches are tailored only for cosegmentation task and do not use localization cues. \\
\par Colocalization is a similar problem~\cite{TJLF14} where the aim is to localize the common object, given a set of images. It was proposed under different names before. For example, object co-detection~\cite{BXS12} is similar, but is given additional bounding boxes and correspondence annotations. Deselaers {\em et al.}~\cite{DAF12} generated candidate bounding boxes and tried to select the correct box within each image using a conditional random field. Cho {\em et al.}~\cite{cho15}, in contrast, localizes the common object by matching common object parts. However, all these approaches are designed for colocalization alone.

\par Our work is mainly inspired by the discriminative framework, proposed first for cosegmentation in Joulin {\em et al.}~\cite{JBP10} and later extended for colocalization by Tang {\em et al.}~\cite{TJLF14} \& Joulin {\em et al.} ~\cite{Joulin14}. We first briefly explain the two main components of the discriminative framework of ~\cite{JBP10}.\\
\par \noindent \textbf{Discriminative clustering.}
 Xu {\em et al.}~\cite{MaxMar05} first proposed the idea of using supervised classifier such as SVM to perform unsupervised clustering. It formulates the clustering problem into  the following optimization problem :
\begin{equation}
\min_{\vect{y}\in\{0,1\}^n,\vect{\alpha}\in\RR^d} 
l(\vect{y},\mat{X}\vect{\alpha}+b\vect{1})+\beta||\vect{\alpha}||^2,
\label{diffrac}
\end{equation}
where $\mat{X}$ is an $n\times d$ feature matrix (also known as design matrix),
$l:\RR^d\rightarrow \RR$ is some loss function, and
$\vect{\alpha}$ a weight vector in $\RR^d$ and scalar $b$ are the parameters of a linear classifier. When
$l$ is the square loss, ~\cite{BaHa07} shows that the problem is equivalent to
\begin{equation}
\min_{\vect{y}\in\{0,1\}^n}\vect{y}^T \mat{D}
\vect{y},
\label{eq:diffrac2}
\end{equation} 
where
\begin{equation}
\mat{D}=\Pi[\mat{I}_d-\mat{X}(\mat{X}^T\Pi\mat{X}+\beta\mat{I}_d)^{-1}\mat{X}^T]
\Pi,\end{equation}
Note that $\mat{I}_d$ is an identity matrix of dimension $d$, $\Pi=\mat{I}_d-\frac{1}{n}\vect{1}\vect{1}^T$ is the usual centering projection matrix and $\mat{D}$ is positive semi-definite. We refer to ~\cite{BaHa07} for more details. \\
\par \noindent \textbf{Local Spatial Similarity} To enforce spatial consistency, a similarity term is combined with the discriminative term $\vect{y}^T \mat{D}
\vect{y}$. The similarity term $\vect{y}^T\mat{L}\vect{y}$ is based on the idea of normalised cut~\cite{ShiMa00} that encourages nearby superpixels with similar appearance to have the same label. Thus, a similarity matrix $\mat{W}^i$ is defined to represent local interactions between superpixels of same image. For any pair of $(a,b)$ of superpixels in image $i$ and for positions $p_a$ and color vectors $c_a$, :
\begin{equation*}
\mat{W}_{ab}^i = \exp(-\lambda_p||p_a-p_b||_2^2 - \lambda_c||c_a-c_b||^2)
\end{equation*}
The $\lambda_p$ is set empirically to .001 \& $\lambda_c$ to .05. Normalised laplacian matrix is given by:
\begin{equation}
\mat{L} = \mat{I}_N- \mat{Q}^{-1/2}\mat{W} \mat{Q}^{-1/2} 
\end{equation} where $\mat{I}_N$ is an identity matrix of dimension $d$, $\mat{Q}$ is the corresponding diagonal {\em degree matrix}, with $Q_{ii}=\sum_{j=1}^n  w_{ij}$.\\

\par  Rest of the paper is organized as follows: Section $3$ describes our novel joint framework. Section 4 gives the implementation details while section 5 evaluates it for the task of cosegmentation on three benchmark datasets. We then move on to colocalization experiments. Lastly, we conclude with discussions of empirical and qualitative results.
 
\section{Joint Colocalization \& Cosegmentation}
\noindent\textbf{Notation.} We use italic Roman or Greek letters (e.g., $x$
or $\gamma$) for scalars, bold italic fonts (e.g.,
$\vect{y}=(y_1,\ldots,y_n)^T$) for vectors, and calligraphic ones
(e.g., $\mat{C}$) for matrices.  We assume we have 
$m$ bounding boxes per image. 

\subsection{Formulation for one Image}
 For the sake of simplicity and clarity, let us first consider a single image, and a set of $m$ bounding boxes per image, with a binary vector $\vect{z}$ in $\{0,1\}^m$ such that
$z_i=1$ when bounding box $i$ in $\{1,\ldots,m\}$ is in the foreground
and $z_i=0$ otherwise. We oversegment the image into $n$ superpixels and define the global superpixel binary vector $\vect{y}$ in $\{0,1\}^n$ such that $y_j=1$ when superpixel number $j$ in $\{1,\ldots,n\}$ is in the foreground and $y_j=0$ otherwise. We also
compute a normalized saliency map $M$ (with values in [0, 1]), and define : $\vect{s} = -log(M)$.\\
\par \noindent\textbf{An Image as a collection of bounding boxes.} We define our optimization problem (in particular, linear constrains) over bounding box and superpixel level. This requires an additional indexing of superpixels on bounding box level and thus, we maintain the following encoding of superpixels: for each bounding box, we maintain the set $S_i$ of its superpixels and  define the corresponding indicator vector $\vect{x}_i$ in $\{0,1\}^{|S_i|}$ such that $x_{ij}=1$ when superpixel $j$ of bounding box $i$ is in the foreground, and $x_{ij}=0$ otherwise. 
Note that  $\vect{x}$ (indexing at bounding box level)  and $\vect{y}$ (indexing at image level) are related by a linear constraint. We  define an indicator projection matrix $\mat{P}$ that encodes the occurrence of a superpixel in all bounding box by $1$ and $0$ as follows: for every box $i$, we define a matrix $\mat{P}_i$ of dimensions $|S_i|$ $\times$ $n$ such that $P_{ij}$ is 1 if superpixel $j$ is present in bounding box $i$ and $0$ otherwise.\\
  
\par \noindent\textbf{Optimization Problem.}
We propose to combine the objective function defined for cosegmentation and colocalization and thus, define:
\begin{equation}
\label{eq:func}
E(\vect{y},\vect{z})=\vect{y}^T(\mat{D}_s+\alpha\mat{L}_s)\vect{y}
+\vect{z}^T\mat{D}_b\vect{z} + \nu\vect{y}^T\vect{s_s} + \mu\vect{z}^T\vect{s_b},
\end{equation}
 The quadratic term $\vect{z}^T\mat{D}_b\vect{z}$ penalizes the selection of bounding boxes whose features are not easily linearly separable from the other boxes. Similarly, minimizing $\vect{y}^T\mat{D}_s\vect{y}$ encourages the most discriminative superpixels to be in the foreground.  Minimizing the similarity term $\vect{y}^T\mat{L}_s\vect{y}$ encourages nearby similar superpixels to have same label whereas the linear terms $\vect{y}^T\vect{s_s}$ and $\vect{z}^T\vect{s_b}$ encourage  selection of salient superpixels and bounding box respectively. Given the feature matrix for superpixels and bounding box, the matrix $\mat{D}_s$ and $\mat{D}_b$ are computed by Equation $3$ whereas $\mat{L}_s$ is computed by Eq.$4$. We define the features and value of scalars later in the implementation detail.\\
\par We now impose appropriate constraints and define the optimization problem as follows:

\begin{equation*}
 \min_{\vect{y},\vect{z}} \quad  E(\vect{y},\vect{z})\quad
\text{under the constraints:}\label{eq:full1}
\end{equation*}
\begin{align}
\gamma |S_i| z_i  \le \sum_{j\in S_i} x_{ij} &\le  (1-\gamma)|S_i| z_i
 & \text{for}\quad i =1,\ldots,m,\label{eq:cons1}\\
\sum_{i:j\in S_i} x_{ij} &\le  \sum_{i:j\in S_i}
z_i,  & \text{for}\quad j=1,\ldots,n,\label{eq:cons2}\\
 \mat{P}_i \hspace{.05in}\vect{y}  &=  \vect{x_i} ,  &\text{for}\quad i=1,\ldots,m. \label{eq:cons4} \\
\sum_{i=1}^m z_i  &= 1  \label{eq:cons3}
\end{align}
The constraint (\ref{eq:cons1}) guarantees that when a bounding box is in the
background, so are all its superpixels, and when it is in the foreground, a
proportion of at least $\gamma$ and at most (1-$\gamma$ of its superpixels are
in the foreground as well, with $0\le\gamma\le 1$. We set $\gamma$ to .1. The constraint (\ref{eq:cons2}) guarantees that a superpixel is in the foreground for only
 one box, the foreground box that contains it (only one of the variables $z_i$ in the summation can be equal to 1). For each bounding box $i$, the constraint (\ref{eq:cons4}) relates the two indexing of superpixels, $\vect{x}$ and $\vect{y}$, by a projection matrix $\mat{P}_i$ defined earlier. The constraint (\ref{eq:cons3}) guarantees that there is exactly one foreground box per image. 
We illustrate the above optimization problem by a toy example of 1 image and 2 bounding boxes in appendix at the end.\\
\par In equations (\ref{eq:func})-(\ref{eq:cons3}), we obtain an integer quadratic program. Thus, we relax the boolean constraints, allowing $\vect{y}$ and $\vect{z}$ to take any value between 0 and 1. The optimization problem becomes convex since all the matrix defined in equation(\ref{eq:func})are positive semi-definite~\cite{JBP10} and the constraints are linear. Given the solution to the quadratic program, we obtain the bounding box by choosing $z_i$ with highest value . For superpixels, since the value of $x$ (and thus $y$) are upper bounded by $z$, we first normalize  $y$ and then, round the values to $0$ (background) and $1$ (foreground).\\

\par \noindent \textbf{Why Joint Optimization.} We briefly visit the intuition behind joint optimization. Note that the superpixel variables $\vect{x}$ and $\vect{y}$ are bounded by bounding box variable $\vect{z}$ in Eq. $6$ and $7$. If the discriminative colocalization part considers some bounding box $z_i$ to be background and sets it to close to 0, this , in principle, enforces the cosegmentation part that superpixels in this bounding box are more likely to be background ($=0$)as defined by the right hand side of Equation $6$: $\sum_{j\in S_i} x_{ij} \le  \delta|S_i| z_i$. Similarly, the segmentation cues influence the final score of $z_i$ variable if the superpixels inside this bounding box are highly discriminative and more likely to be foreground.
\section{Implementation Details}
  We will release source code of our implementation at the time of publication. We use superpixels obtained from publicly available implementation of~\cite{QS08}. This reduces the size of the matrix $\mat{D}_s$,$\mat{L}_s$ and allows us to optimize at superpixel level. Using the publicly available implementation of~\cite{AlexFer12}, we generate 20 bounding boxes for each image. We use unsupervised method of ~\cite{YXSJ13} to compute off the shelf saliency maps in our experiments.\\
\par \noindent \textbf{Features.} Following ~\cite{JBP10}, we densely extract SIFT features at every 4 pixels and  kernelize them using Chi-square distance. For each bounding box, we extract 4096 dimensional feature vector using AlexNet~\cite{Alex12}. \\
\par \noindent \textbf{Hyperparameters}
Following ~\cite{TJLF14}, we set $\mu$, the balancing scalar for box saliency, to $.001$. To set $\alpha$, we follow ~\cite{JBP10} and set it $\alpha=.1$ for foreground objects with fairly uniform colors, and  $= .001$ corresponding to objects with sharp color variations. We empirically set scalar $\nu =.005$ by optimizing over a small set.\\

\section{Evaluation of Joint Framework}
The goal of this section is two fold: First, we propose several baselines that help understand the individual contribution of various cues in the optimization problem defined in section $3.1$. Second, we empirically validate and show that learning the two problems jointly significantly improve the performance over learning them individually. 
\subsection{Cosegmentation Experiments}

\subsubsection{Baseline Methods}
 In the section 3.1, we make the following two changes to the cosegmentation framework of ~\cite{JBP10}: First, we add the saliency cues as a linear term $\vect{s_p}$ to the framework of ~\cite{JBP10}. Second, we propose to optimize the objective function of ~\cite{JBP10}  with a quadratic program(QP) solver whereas in ~\cite{JBP10}, it is optimized with a semi-definite programming (SDP) solver~\cite{JournBa08}. To illustrate the importance of saliency cues and better understand the different optimization techniques, we propose the following baseline methods: 
\par \noindent{\textbf{B1.}}  Discriminative clustering~\cite{BaHa07} objective is  usually optimized with a SDP solver as the semi-definite relaxations are strong and do not suffer from trivial solutions. To validate this, we optimize  the objective function of ~\cite{JBP10} with a quadratic program (QP)  solver and compare the results with the SDP solver of ~\cite{JournBa08}.\\ 
 \par \noindent{\textbf{B2.}} To quantify the impact of saliency cues, we propose to solve a linear program that obtains an image segmentation by finding the most salient pixels. Thus, we minimize a linear saliency term $\vect{s_s}$ under a linear constraint that minimum number of foreground pixels should be greater than a fraction of total image pixels. This basically means choosing a fraction of the most salient pixels in an image. We set the fraction to .4 as a rough measure of total foreground pixels in MSRC~\cite{Shotton06} and Object Discovery dataset~\cite{RJKL13}.\\
\par \noindent{\textbf{B3.}} To illustrate the benefits of combining discriminative framework and saliency cues, we solve a QP that optimizes the new objective function of ~\cite{JBP10} that includes the saliency cues.\\

\par \noindent We denote the results obtained from our joint framework by Ours. In addition to the baselines proposed above and JBP10~\cite{JBP10}, we compare our method with four state of the art approaches RJKL13~\cite{RJKL13},WHG13 ~\cite{WHG13}, FI13~\cite{Irani13} and QHZN16~\cite{Quan_16}. Unless stated otherwise,  we measure the segmentation accuracy as the percentage of pixels labeled accurately i.e. average precision (AP).  
\subsubsection{Benchmark Datasets.}
We evaluate the cosegmentation performance of our framework on three benchmark  datasets: MSRC~\cite{Shotton06}, Object Discovery dataset~\cite{RJKL13} and PASCAL-VOC 2010. MSRC contains a subset of 10 object classes, each containing 24 to 30 images. The Object Discovery dataset ~\cite{RJKL13} was collected by downloading images from Internet for airplane, car and horse. It is significantly larger and thus, diverse in terms of viewpoints, texture, color etc. Faktor \& Irani~\cite{Irani13} collected a subset of PASCAL-VOC 2010 dataset to evaluate the cosegmentation performance. This subset is obtained by choosing images in which the total size of a co-object is at least 1\% of the image size. Overall, it contains 1037 images from the 20 PASCAL classes. \\
\par In Table $1$, $2$ and $3$, we show our results and comparison with other approaches on these three datasets. Note that the results mentioned for JBP10~\cite{JBP10} are obtained by running their open source code and verified with the authors while for others, we simply cite their numbers from their paper. WHG13~\cite{WHG13} shows results on MSRC in two modes: supervised and unsupervised. We compare with unsupervised performance on six classes from their paper. For fair comparison with the state of the art~\cite{Quan_16} on Object Discovery dataset and PASCAL-VOC 2010, we report performance obtained by applying grab cut~\cite{GrabCut} based post processing on our output.\\  
 
\begin{table}[t]
\caption{Comparison  on Object Discovery dataset.}
\label{seg-obj-table}
 \centering
\scalebox{0.9}{\begin{tabular}{ l|lllllll} 
\toprule
 Class  & JBP10 & B1 & B2 & B3 & Ours & RJKL13. & QHZN16 \\
 \midrule
  Horse  & 69 & 64 &70 &  72& 87 &82.8 & 89.3   \\ 
 Plane  & 65 & 71 &64 & 71&  86&85.3  & 91.0 \\
 Car  & 75 & 63 & 79 & 79& 87&82.0 & 88.5 \\
\midrule
 Avg. & 69.7 & 66 & 71.3 & 74.0 & 86.6& 83.4 & 89.6 \\
\bottomrule
\end{tabular}
}
\vskip 0.15in
\caption{ Comparison  on MSRC dataset}
\label{seg-msrc-table}

\scalebox{0.95}{\begin{tabular}{l|lllllll} 
 \toprule
 Class  &JBP10 & B1 & B2 & B3 & Ours & RJKL13 & WHG13  \\
 \midrule
 Dog  & 73 & 62 & 72   &75  & 87  & 92&-\\
 Chair  & 74 & 69 & 73 & 78& 84 & 84&-  \\
 Sheep  & 83 & 73 & 74 & 80& 92 &90 & -\\ 
 Bike  & 64 & 65 & 64 & 66& 76  &78& 74.8  \\
 Plane  & 53 & 73 & 65 & 74& 84 & 82&87.3  \\
 Cow  & 80 & 69 & 76 & 80 & 89  &92 & 89.7 \\
 Car  & 68 & 51 & 75 & 78 & 82 &82& 90.0\\
 Face  & 75 & 62 & 73 & 76& 82  & 82& 89.3\\
 Cat  & 70 & 65 & 72 & 76& 84 & 90& 88.3 \\
 Bird  & 78 & 65 & 73 & 77& 90 & 90& - \\
 \midrule
 Avg. & 71.1 & 65.4 &71.5& 75.9& 85.0 & 86.2 & 86.6\\
  \bottomrule
\end{tabular}
}
\end{table}

\begin{figure*}[t]
 \vskip .1in 
        \includegraphics[height=0.15\linewidth,width=.19\linewidth]{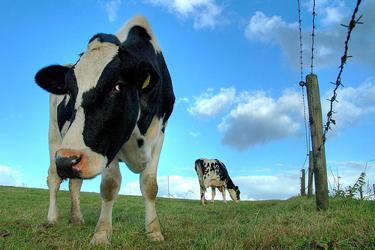}            
        \includegraphics[height=0.15\linewidth,width=.19\linewidth]{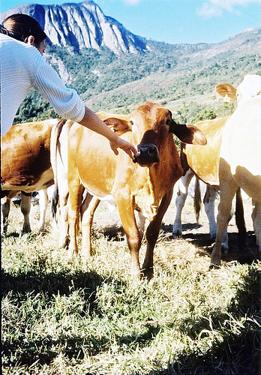}
       \includegraphics[height=0.15\linewidth,width=.19\linewidth]{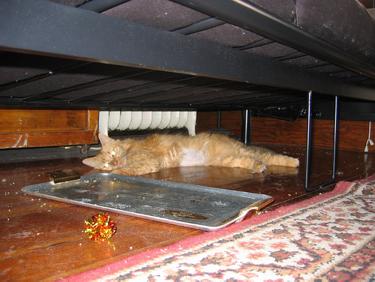}
      \includegraphics[height=0.15\linewidth,width=.19\linewidth]{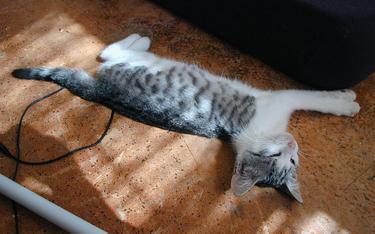}       
        \includegraphics[height=0.15\linewidth,width=.19\linewidth]{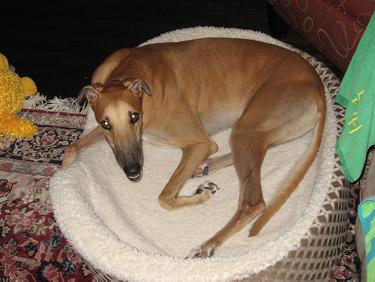}
      \\
      \includegraphics[height=0.15\linewidth,width=.19\linewidth]{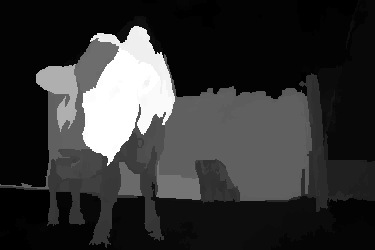}            
        \includegraphics[height=0.15\linewidth,width=.19\linewidth]{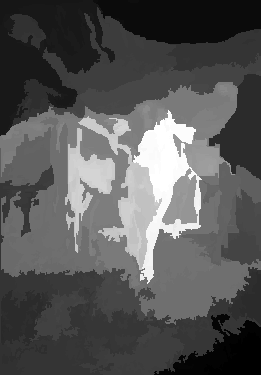}
       \includegraphics[height=0.15\linewidth,width=.19\linewidth]{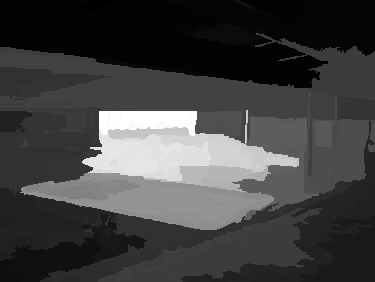}
      \includegraphics[height=0.15\linewidth,width=.19\linewidth]{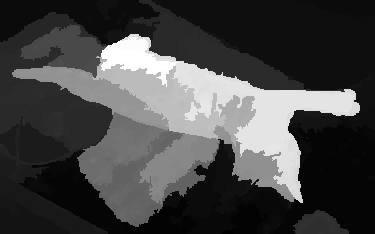}       
        \includegraphics[height=0.15\linewidth,width=.19\linewidth]{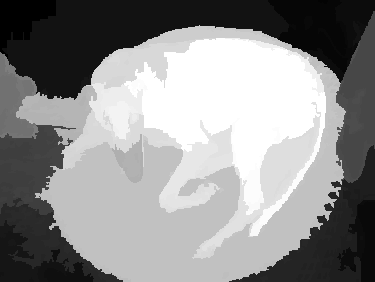}
      \\
     \includegraphics[height=0.15\linewidth,width=.19\linewidth]{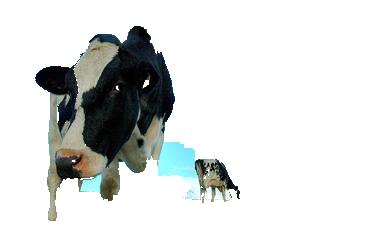}            
        \includegraphics[height=0.15\linewidth,width=.19\linewidth]{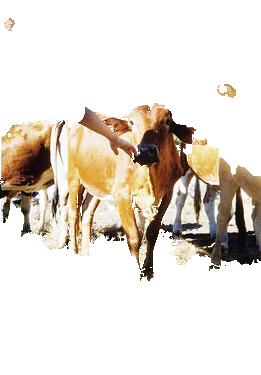}
       \includegraphics[height=0.15\linewidth,width=.19\linewidth]{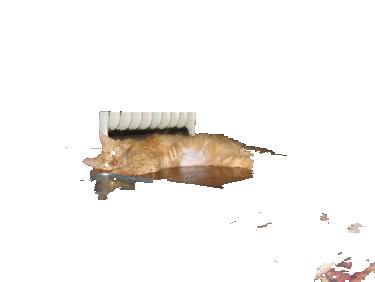}
      \includegraphics[height=0.15\linewidth,width=.19\linewidth]{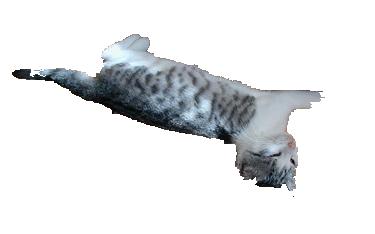}       
        \includegraphics[height=0.15\linewidth,width=.19\linewidth]{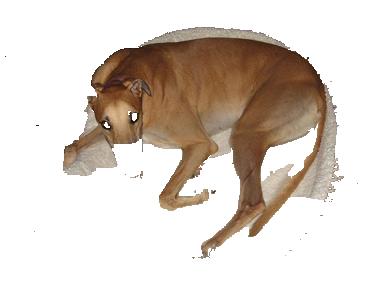}
    
\caption{Qualitative results on challenging Pascal VOC 2010 images. Top row contains input images, middle row depicts the saliency map and bottom row shows the segmented foreground.} 
\vskip .1in
\end{figure*}
\par \noindent \textbf{MSRC Dataset} In Table $2$, we observe that the B1 is consistently outperformed by the SDP solver of JBP10~\cite{JBP10} on both datasets by an average margin of 5 \% AP. However, B3 consistently improves the performance of JBP10~\cite{JBP10} by an average of 5 \% AP. This shows that the objective function of ~\cite{JBP10}, combined with saliency 
cues, can be optimized efficiently and accurately with a QP solver. Also, only saliency based segmentation, B2,  gives a reasonable accuracy of 71\% AP. Compared to JBP10~\cite{JBP10}, our framework improves the average precision on MSRC dataset by almost $14$ \%. Our results compete well with RJKL13~\cite{RJKL13}, on 6 out of 10 classes on MSRC dataset. \\
\par \noindent \textbf{Experiments on Object Discovery Dataset} In Table $1$, we observe the same trend. We improve upon the result of JBP10~\cite{JBP10} by $14$ \% and consistent gains over the baselines demonstrate the robustness of the model using localization cues. We outperform RJKL13 on all three classes.  We compare well with QHZN16~\cite{Quan_16} on classes Car and Horse but worse on aeroplane class.  \\

\par \noindent \textbf{Experiments on Pascal VOC 2010} As argued by Faktor \& Irani~\cite{Irani13}, average precision metric is not reliable to evaluate cosegmentation algorithm on this subset as  90 \% of overall image content lies in background. Therefore, in addition to average precision(AP), we also evaluate our algorithm using  Intersection over union (IoU) metric, also known as Jaccard similarity, in Table $3$. We compare with two state of the art approaches that have shown results on this dataset yet: FI13 ~\cite{Irani13} and QHZN16. We outperform FI13~\cite{Irani13} in both metrics but perform sightly worse than QHZN16~\cite{Quan_16}.

\begin{table}[h]
 \centering
\caption{Comparison on Pascal VOC2010}
\vskip .1in
\begin{tabular}{ |l|l|l|l|} 
\toprule
  Metric  & Ours & FI13 & QHZN16 \\
 \midrule
  Mean IoU   & 47& 46  &52 \\ 
  \midrule
  AP   & 86 & 84  & 89\\ 
\bottomrule
\end{tabular}
\end{table}

 \par \noindent \textbf{Qualitative Results}
 In Figure $2$ and $3$, we provide some examples of our end result. Figure $2$ illustrates that our framework considers saliency as one of the various helpful cues and is robust to not salient objects or incorrect saliency maps. For example, in the very first example in Figure $2$, the smaller cow is not at all salient  and yet, our method rightly segments it as a foreground. In Figure $3$, we show some examples from MSRC dataset. on top, we have an original image shown with the selected bounding box and underneath, we show the segmentation of the whole image. 
 
\subsection{Colocalization Experiments}
\par \noindent \textbf{Evaluation Metrics.} We conduct colocalization experiments on PASCAL VOC 2007~\cite{voc-07}. We use two evaluation metrics to compare with state of the art colocalization techniques: \par1) The standard Intersection over union (IoU) metric for object detection(intersection of predicted bounding box area and ground-truth bounding box area divided by the area of their union)
\par 2) Correct  Localization (CorLoc)  metric,  an  evaluation  metric  used  in  related  work ~\cite{TJLF14,cho15}, and defined as the percentage of images correctly localized according to the criterion: $IoU > .5$. 

\subsubsection{Baseline Methods}
 We analyze individual components of our colocalization
model by removing various terms in the objective function and consider the following 
baselines:
\par \noindent \textbf{Sal.} This baseline only minimizes the saliency term for bounding boxes, without any segmentation cue, and picks the most salient one in each image. It is important as it gives an approximate idea about which object classes are more salient in the dataset.\\

\newlength{\imagewidth}
\settowidth{\imagewidth}{\includegraphics{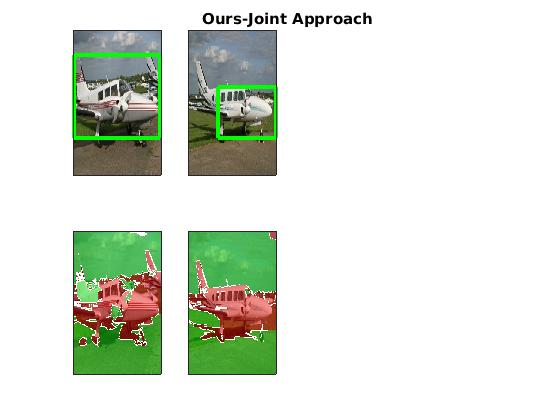}}
 \begin{figure*}[t]
 \vskip .15in
    \begin{minipage}{.25\linewidth}
        \centering
        \includegraphics[trim=0.05\imagewidth{} 0.05\imagewidth{} 0.5\imagewidth{} 0.05\imagewidth{}, clip, width = .99\linewidth]{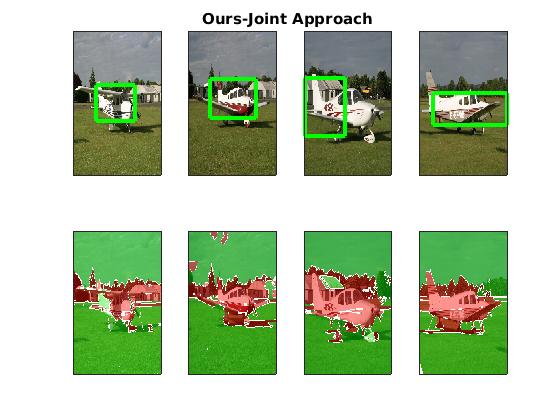}
    \end{minipage}%
    \begin{minipage}{.25\linewidth}
        \centering
        \includegraphics[trim=0.5\imagewidth{} 0.05\imagewidth{} 0.05\imagewidth{} 0.05\imagewidth{}, clip, width = .99\linewidth]{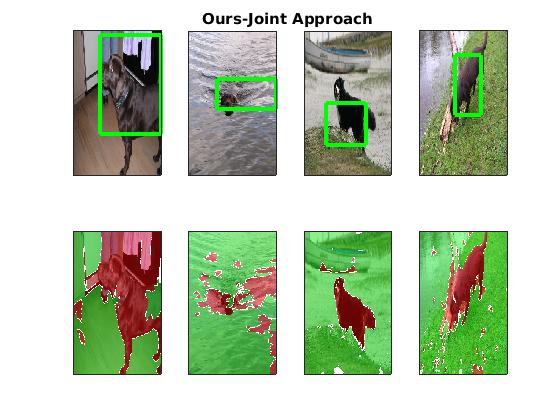}
    \end{minipage}%
    \begin{minipage}{.25\linewidth}
        \centering
        \includegraphics[trim=0.5\imagewidth{} 0.05\imagewidth{} 0.05\imagewidth{} 0.05\imagewidth{}, clip, width = .99\linewidth]{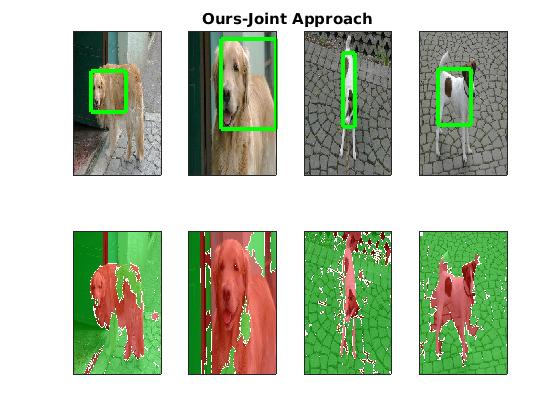}
    \end{minipage}%
    \begin{minipage}{.25\linewidth}
        \centering
        \includegraphics[trim=0.05\imagewidth{} 0.05\imagewidth{} 0.5\imagewidth{} 0.05\imagewidth{}, clip, width = .99\linewidth]{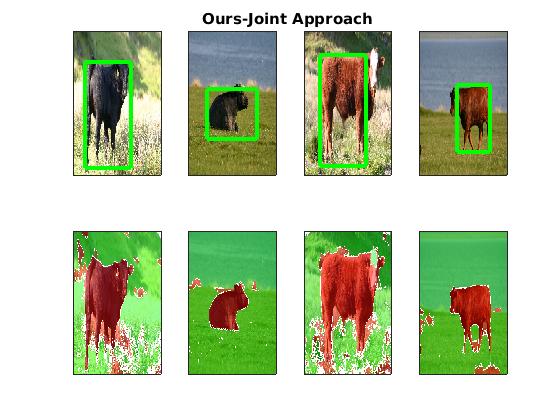}
    \end{minipage}\\
\caption{Examples of joint colocalization and cosegmentation on MSRC dataset} 
\vskip .15in
      \end{figure*}
  
\par \noindent \textbf{Sal+Disc.} This baseline includes the saliency and discriminative term for boxes, without any segmentation cues. \\

\par \noindent \textbf{TJLF14} Tang {\em et al.},TJLF14 ~\cite{TJLF14} tackles colocalization alone without any segmentation spatial support. It quantifies how much we gain in colocalization performance by leveraging segmentation cues. 

\begin{table}[t]
\centering
\caption{Comparison on Pascal VOC-2007}
\vskip .15in
 \begin{tabular}{l|lllll}
\toprule
 Class  &Sal & Sal+Disc & TJLF14 & Ours  \\
 \midrule
 Plane-Left  & 32 & 41  & 42& 42\\
 Plane-Right  & 20&43  & 51&59 \\
 Boat-Left  &  02& 06  &11 & 16 \\ 
 Boat-Right  &09& 09 & 12 & 21   \\
 Bike-Left  & 48& 30 & 51& 41 \\
 Bike-Right  &  47& 41 &65 & 56  \\
 Horse-Left  & 31 & 25 & 44 & 38\\
 Horse-Right  & 36  & 34 & 52 &52\\
 Bus-Left  & 14 & 14 & 38 & 53  \\
 Bus-Right  & 39& 39 &  57& 65 \\
 Bicycle-Left & 25 & 27 & 25 & 35\\
 Bicycle-Right & 28 & 32 & 24 & 45\\
 \midrule
 Mean CorLoc. &28 & 29 & 39 & 44 \\
  \bottomrule
\end{tabular}
\end{table} 

\subsubsection{Colocalization evaluation on Pascal VOC 2007 } 
Following the experimental setup defined in ~\cite{DAF12,TJLF14,cho15}, we evaluate our method on the PASCAL07-6x2 subset to compare to previous methods for co-localization.  This subset consists of all images from 6 classes (aeroplane, bicycle, boat,
bus, horse, and motorbike) of the PASCAL VOC 2007~\cite{voc-07}.  Each  of  the 12 class/viewpoint combinations contains between 21 and 50 images for a total of 463 images. Compared  to  the  Object  Discovery dataset, it is significantly more challenging due to considerable clutter, occlusion, and diverse viewpoints.\\

\par \noindent \textbf{CorLoc Metric} 
In Table $4$, we report our experiments based on CorLoc Metric and compare them with our baselines proposed before. Note that we use plane instead of Aeroplane and bike instead of Motorbike. We see that results using stripped down versions of our model are not consistent and less reliable. In particular, we observe that on salient classes, saliency term alone, $Sal.$, is a very strong baseline. This can also be partially predicted by looking at the individual images of Bike class. However, it fails badly on classes such as Boat and Bus which are more cluttered, occluded and exhibit huge change in scale space. $Sal+Disc$ improves upon the saliency baseline only in some classes such as Plane but overall fails to improve upon TJLF14 without segmentation cues. Our results improve upon both $Sal.$ and $Sal+Disc$ in all classes. This again validates our hypothesis of leveraging segmentation cues to lift the colocalization performance. Our results outperforms TJLF14~\cite{TJLF14} on most classes. \\
\par \noindent \textbf{IoU Metric.}
In Figure $4$, we show failure cases of our colocalization results based on CorLoc Metric. In the first row, we show several instances where the localization is near perfect and yet, the bounding box only achieves the CorLoc score of approximately $.45.$ to $.49$ and thus, counts as a failure case according to CorLoc metric. This is mainly because it does not include the tail or a wing of aeroplane inside bounding box. We observe similar cases in class Horse too. To further support our argument quantitatively, we compare our results based on IoU metric with ~\cite{cho15} on Pascal VOC 2007 in Table $5$. We could not compare with TJLF14 on IoU metric as their source code and hyper-parameters are not publicly available. IoU metric gives a value of $.45$ to instances such as shown in Figure $3$ whereas CorLoc gives it a score of 0.

\begin{table}[h]
 \centering
\caption{IoU score on Pascal VOC2007}
\vskip .1in
\begin{tabular}{ |l|l|l|} 
\toprule
  Metric  & Ours & CSP15 \\
 \midrule
  Mean IoU   & 45& 56  \\ 
\bottomrule
\end{tabular}
\end{table}
\par \noindent \textbf{Comparison to state of the art.} Cho {\em et al.}, CSP15~\cite{cho15}, outperforms all approaches by a huge margin on CorLoc metric where it obtains an absolute score of 64. This is partially because it leverages part based matching by Hough Transform where the predicted bounding box is selected by a heuristic standout score. In contrast, the discriminative framework of ours does not incorporate any constraints on including parts of objects in the predicted bounding box. This is also partially evident in Table $5$  where the margin between our performance and CSP15 on IoU metric is almost half that of CorLoc. Moreover, CSP15, by design, is tailored for colocalization only whereas our framework tackles both colocalization and cosegmentation. 
\begin{figure*}[ht]
 \vskip .15in
    \begin{minipage}{.25\linewidth}
        \centering
        \includegraphics[trim=0.02\imagewidth{} 0.02\imagewidth{} 0.02\imagewidth{} 0.02\imagewidth{}, clip, width = .95\linewidth]{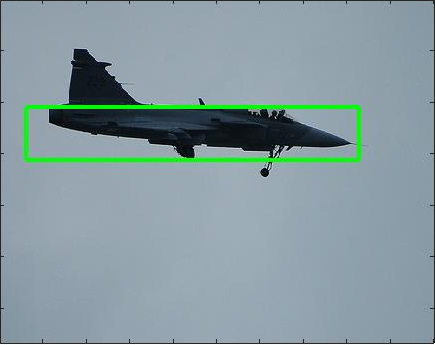}
    \end{minipage}%
    \begin{minipage}{.25\linewidth}
        \centering
        \includegraphics[trim=0.02\imagewidth{} 0.02\imagewidth{} 0.02\imagewidth{} 0.02\imagewidth{}, clip, width = .95\linewidth]{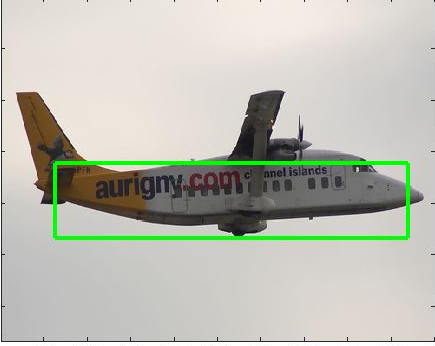}
    \end{minipage}%
    \begin{minipage}{.25\linewidth}
        \centering
        \includegraphics[trim=0.02\imagewidth{} 0.02\imagewidth{} 0.02\imagewidth{} 0.02\imagewidth{}, clip, width = .95\linewidth]{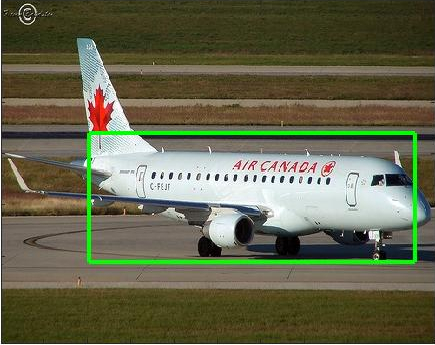}
    \end{minipage}%
      \begin{minipage}{.25\linewidth}
        \centering
        \includegraphics[trim=0.02\imagewidth{} 0.02\imagewidth{} 0.02\imagewidth{} 0.05\imagewidth{}, clip, width = .95\linewidth]{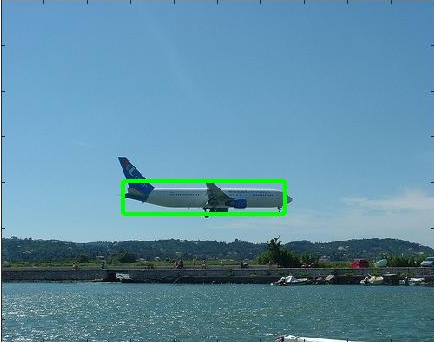}
    \end{minipage} 
    \caption{Some Failure Cases of colocalization according to CorLoc Metric}
    \vskip .15in
\end{figure*}

\section{Conclusion \& Future Work}
We proposed a novel framework that jointly learns to localize and segment objects. The proposed formulation is based on two different level of visual representations and uses linear constraints as a means to transfer information implicit in these representations in an unsupervised manner. Although we demonstrate the effectiveness of our approach with a variant of maximum margin clustering, the key idea of transferring knowledge between tasks at different granularity is general and can be incorporated in the framework of constrained CNN~\cite{ccnn}. Future work could also extend our model by including an image-video classifier, thereby providing a single framework that simultaneously classify, localize and segment common objects or actions in images and videos respectively.
\section{Acknowledgement} This work started as a Master's thesis project at INRIA Willow team and was partially supported by ERC Advanced grants \\VideoWorld and Allegro. Many thanks to Armand Joulin for numerous helpful discussions and comments on this paper. The author also thanks anonymous reviewers for their comments. 
\section{Appendix}
We illustrate our joint colocalization and cosegmentation framework by a simple toy example. Suppose the image contains 5 superpixels. Thus, the  global image level superpixel indexing is defined by $\vect{y}=(y_1,y_2,y_3, y_4,y_5)^T$. Also, assume that there are two bounding boxes per image and that bounding box $1$, $z_1$, contains superpixel $1,3,4 $ while bounding box $2$, $z_2$, contains  superpixel $1,2,4$. Thus, bounding box indexing for first proposal $z_1$ is defined by $\vect{x_1} = (y_1, y_3, y_4) ^T $ and for $z_2$ is defined by $\vect{x_2} = (y_1, y_2, y_4)^T$. Vector $\vect{x}$ is obtained by concatenating $\vect{x_1}$ and $\vect{x_2}$. Then, vector $\vect{x_1}$ and vector $\vect{y}$ are related by an indicator projection matrix $\mat{P}_1$ as follows:
\begin{equation*}
\mbox{\small $\left[ \begin{array}{c} \vect{x_1} \end{array} \right] = \left[ \begin{array}{c} y_1 \\ y_3 \\ y_4 \end{array} \right] =
\underbrace{\begin{bmatrix}
1 & 0 & 0  & 0  & 0 \\ 
0 & 0 & 1  & 0  & 0 \\
0 & 0 & 0  & 1  & 0 \\
\end{bmatrix}}_\text{$\mat{P}_1$} \times \underbrace{\left[\begin{array}{c} y_1 \\ y_2 \\ y_3 \\ y_4 \\ y_5  \end{array} \right]}_\text{\vect{y}} $}
\end{equation*}

 Note that matrix $\mat{P}$ basically tells how many times a superpixel occurs in all bounding boxes or equivalently, how many times $y_i$ is duplicated in the vector $\vect{x}$. 
 We now translate the other three constraints from the paper one by one. Note that $|S_i| = 3$ since each bounding box contains $3$ superpixels, $m = 2$ and $n=5$. To keep it short, we only demonstrate the constraints for the superpixels of the first bounding box($i=1$). 
\begin{align}\nonumber
 &\gamma |S_i| z_i  \le \sum_{j\in S_i} x_{ij} \le  (1-\gamma)|S_i| z_i
   \hspace{.5cm}\text{for}\quad i =1 \\\nonumber
 \Rightarrow &\gamma*3z_1 \le (x_{11} +x_{12} +x_{13}) \le (1-\gamma)*3z_1 \\\nonumber
 \\\nonumber
 \Rightarrow &\gamma*3z_1 \le (y_{1} +y_{3} +y_{4}) \le (1-\gamma)*3z_1 \hspace{.1cm} \text{(By $\mat{P}_1\vect{y} =\vect{x}_1$)}
 \end{align}
Similarly, the second constraint for superpixels is equivalent to:
\begin{align}\nonumber
    \sum_{i:j\in S_i} x_{ij} &\le  \sum_{i:j\in S_i}
z_i,  \text{for}\quad j=1,2,3,4,5 \\\nonumber
\\\nonumber
 (x_{11}+x_{21}) &\le (z_1 +z_2) \Rightarrow 2y_1 \le (z_1 +z_2)\\\nonumber
\\\nonumber
x_{22} &\le z_2  \Rightarrow y_2 \le z_2 \\\nonumber
\\\nonumber
x_{12} &\le z_1  \Rightarrow y_3 \le z_1 \\\nonumber
\\\nonumber
(x_{13} + x_{23}) &\le(z_1 +z_2)  \Rightarrow  2y_4 \le (z_1 +z_2)
\end{align}
Finally, for the bounding boxes, we have:
\begin{align}\nonumber
\sum_{i=1}^m z_i  = 1  
\Rightarrow z_1 +z_2 = 1
\end{align}

\begin{figure*}[t]

        \includegraphics[height=0.17\linewidth,width=.135\linewidth]{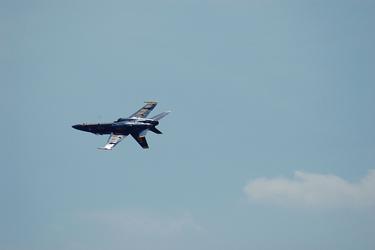} 
        \includegraphics[height=0.17\linewidth,width=.135\linewidth]{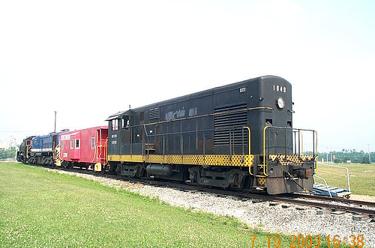} 
        \includegraphics[height=0.17\linewidth,width=.135\linewidth]{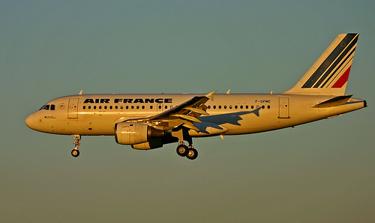}
       \includegraphics[height=0.17\linewidth,width=.135\linewidth]{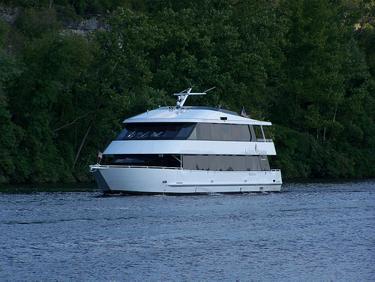}
       \includegraphics[height=0.17\linewidth,width=.135\linewidth]{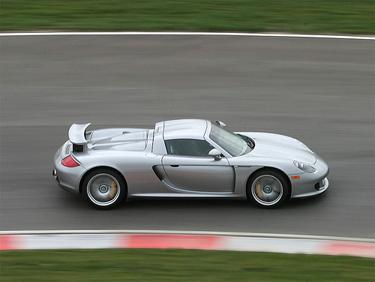} 
      \includegraphics[height=0.17\linewidth,width=.135\linewidth]{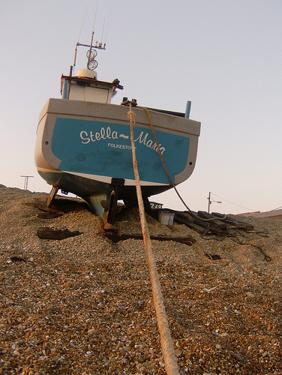}
      \includegraphics[height=0.17\linewidth,width=.135\linewidth]{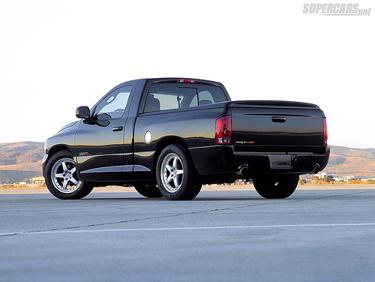} 
      \\
      \includegraphics[height=0.17\linewidth,width=.135\linewidth]{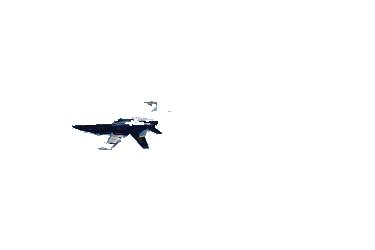} 
      \includegraphics[height=0.17\linewidth,width=.135\linewidth]{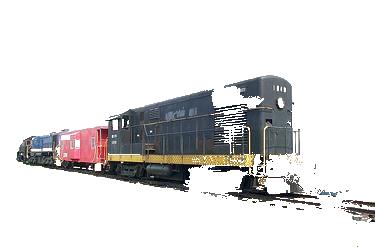}
        \includegraphics[height=0.17\linewidth,width=.135\linewidth]{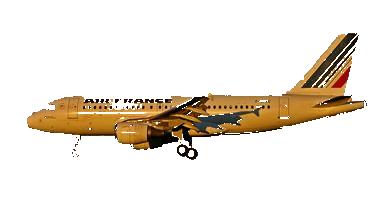}
       \includegraphics[height=0.17\linewidth,width=.135\linewidth]{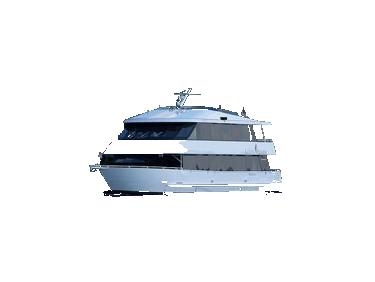}
       \includegraphics[height=0.17\linewidth,width=.135\linewidth]{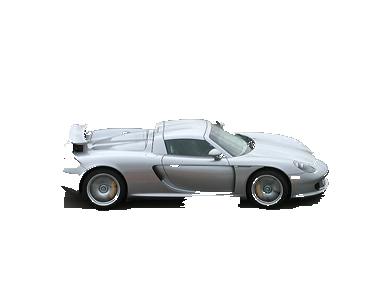} 
      \includegraphics[height=0.17\linewidth,width=.135\linewidth]{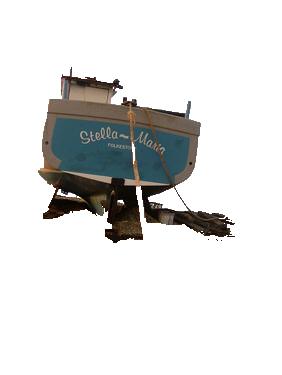} 
       \includegraphics[height=0.17\linewidth,width=.135\linewidth]{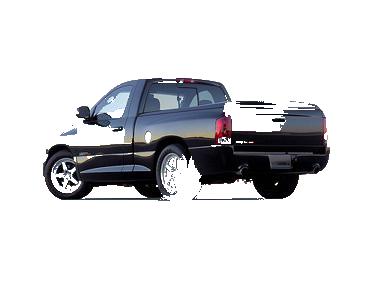}
     
\caption{More Qualitative results .} 
\vskip .1in
\end{figure*}
\bibliography{kreigs}
\bibliographystyle{IEEEtran}
\end{document}